\documentclass{llncs}

\usepackage{times}
\usepackage{latexsym}
\usepackage{alltt}

\newcommand{\eclipse}{ECL$^i$PS$^e$}

\newcommand{\ra}{\mbox{$\:\rightarrow\:$}}

\newcommand{\La}{\mbox{$\:\Leftarrow\:$}}
\newcommand{\Ra}{\mbox{$\:\Rightarrow\:$}}

\newcommand{\sse}{\mbox{$\:\subseteq\:$}}

\newcommand{\LL}{\mbox{$\ldots$}}

\newcommand{\C}[1]{\mbox{$\{{#1}\}$}}           

\newcommand{\NI}{\noindent}
\newcommand{\HB}{\hfill{$\Box$}}
\newcommand{\VV}{\vspace{5 mm}}
\newcommand{\III}{\vspace{3 mm}}
\newcommand{\II}{\vspace{2 mm}}

\newcommand{\szkew}[1]{\relax \setbox0=\hbox{\kern -24pt $\displaystyle#1$\kern 0pt }%
\box0}
{\catcode`\@=11 \global\let\ifjusthvtest@=\iffalse}

\newcounter{oldmycaption}

\newcommand{\p}[2]{\langle #1 \ ; \ #2 \rangle}

\newcommand{\Proof}{\NI
                    {\bf Proof.}\ }
\newcommand{\noprint}[1]{}
\newcommand{\noprintpb}[1]{}

\title{Automatic Generation of Constraint Propagation
Algorithms for Small Finite Domains}

\author{Krzysztof R. Apt\inst{1,2} and Eric Monfroy\inst{1}}
\institute{CWI\\
P.O. Box 94079, 1090 GB Amsterdam, the Netherlands\\
\email{\{K.R.Apt,Eric.Monfroy\}@cwi.nl}\\
\and 
University of Amsterdam, the Netherlands}

\begin{document}

\maketitle
\begin{abstract}

  We study here constraint satisfaction problems that are ba\-sed on
  predefined, explicitly given finite constraints. To solve them we
  propose a notion of {\em rule consistency\/} that can be expressed
  in terms of rules derived from the explicit representation of the
  initial constraints.

  This notion of local consistency is weaker than arc consistency for 
  constraints of arbitrary arity but
  coincides with it when all domains are unary or binary. For Boolean
  constraints rule consistency coincides with the closure under the
  well-known propagation rules for Boolean constraints.

  By generalizing the format of the rules we obtain a characterization
  of arc consistency in terms of so-called inclusion rules.  The
  advantage of rule consistency and this rule based characterization
  of the arc consistency is that the algorithms that enforce both
  notions can be automatically generated, as {\tt CHR} rules. So these
  algorithms could be integrated into constraint logic programming
  systems such as \eclipse{}.

  We illustrate the usefulness of this approach to constraint propagation
  by discussing the implementations of both algorithms and their use
  on various examples, including Boolean constraints, three
  valued logic of Kleene, constraints dealing with Waltz's language
  for describing polyhedreal scenes, and Allen's qualitative approach
  to temporal logic.
\end{abstract}

\section{Introduction}

In constraint programming the programming process is limited to a
generation of constraints and a solution of the so obtained constraint
satisfaction problems (CSP's) by general or domain dependent methods.

On the theoretical side several notions of local consistency, notably
arc consistency for constraints of arbitrary arity, have been defined
and various search methods have been proposed. On the practical side
several constraint programming systems were designed and implemented
that provide a substantial support for constraint programming.  This
support is usually provided in the form of specific built-in
constructs that support search and constraint propagation.  For
example, the arc consistency is built in the ILOG Solver and is
present in a library of the most recent version of \eclipse{}.

In this paper we study CSP's that are built out of predefined,
explicitly given finite constraints.  Such CSP's often arise in
practice. Examples include Boolean constraints, constraints dealing
with Waltz's language for describing polyhedreal scenes, Allen's
temporal logic, and constraints in any multi-valued logic.

In such situations it is natural to explore the structure of these
explicitly given constraints first and to use this information
to reduce the considered CSP to a simpler yet equivalent one.
This information can be expressed in terms of rules.
This leads to a local consistency notion called {\em rule consistency}
that turns out to be weaker than arc consistency for constraints of
arbitrary arity.

When the original domains are all unary or binary, rule
consistency coincides with arc consistency.  When additionally the
predefined constraints are the truth tables of the Boolean
connectives, these rules coincide with the well-known
rules for Boolean constraints, sometimes called unit propagation
rules (see, e.g. \cite{Dal92}). As a
side effect, this shows that the unit propagation rules characterize
arc consistency.  Rule consistency is thus a generalization of
the unit propagation to non-binary domains.

Next, we show that by generalizing the notion of rules to so-called
inclusion rules, we obtain a notion of local consistency that
coincides with arc consistency for constraints of arbitrary arity.

The advantage of the rule consistency and this rule based
characterization of the arc consistency is that the algorithms that
enforce them can be automatically generated and provided on the
constraint programming language level.  For example, the rules in
question can be generated automatically and represented as rules of
the {\tt CHR} language of \cite{fruhwirth-constraint-95} that is part
of the \eclipse{} system.  (For a more recent and more complete
overview of {\tt CHR} see \cite{FruehwirthJLP98}.)

Consequently, the implementations of the algorithms that achieve rule
consistency and arc consistency for the considered CSP's are simply
these automatically generated {\tt CHR} programs.  When combined with
a labeling procedure such {\tt CHR} programs constitute automatically
derived decision procedures for these CSP's.

The availability of the algorithms that enforce rule consistency and
arc consistency on the constraint programming language level further
contributes to the automatization of the programming process within
the constraint programming framework. In fact, in the case of such
CSP's built out of predefined, explicitly given finite constraints the
user does not need to write one's own {\tt CHR} rules 
for the considered constraints and can simply
adopt all or some of the rules that are automatically generated.  In
the final example of the paper we also show how using the rules and
the inclusion rules, we can implement more powerful notions of local
consistency.

Alternatively, the generated rules and inclusion rules could be fed
into any of the generic {\em Chaotic Iteration\/} algorithms of
\cite{Apt99b} and made available in such systems as the ILOG solver.
This would yield rule consistency and an alternative implementation 
of arc consistency.

The algorithms that for an explicitly given finite constraint generate
the appropriate rules that characterize rule consistency and arc
consistency have (unavoidably) a running time that is exponential in
the number of constraint variables and consequently are in general
impractical.

To test the usefulness of these algorithms for small finite domains we
implemented them in \eclipse{} and successfully used them on several
examples including the ones mentioned above. The fact that we could
handle these examples shows that this approach is of practical value
and can be used to automatically derive practical decision procedures
for constraint satisfaction problems defined over small finite
domains.  Also it shows the usefulness of the {\tt CHR} language for
an automatic generation of constraint solvers and of decision procedures.

The rest of the paper is organized as follows. In the next section
we formalize the concept that a CSP is built out of predefined constraints.
Next, in Section
\ref{sec:rule-consistency} we
introduce the notion of a rule, define the notion of rule consistency
and discuss an algorithm that can be used to generate the minimal set
of rules that
characterize this notion of local consistency. Then, in Section
\ref{sec:relation-arc-consistency} we compare rule consistency to arc
consistency. In Section \ref{sec:inclusion-rule-consistency} we
generalize the notion of rules to so-called inclusion rules and
discuss an algorithm analogous to the one of
Section~\ref{sec:rule-consistency}. This
entails a notion of local consistency that turns out to be equivalent
to arc consistency.  Finally, in Section \ref{sec:examples} we discuss
the implementation of both algorithms. They generate from an explicit
representation of a finite constraint a set of {\tt CHR}
rules that characterize respectively rule consistency and arc
consistency. We also illustrate the usefulness of these implementations
by means of several examples.
Due to lack of space all proofs are omitted.

\section{CSP's Built out of Predefined Constraints}
\label{sec:predefined}


Consider a finite sequence of variables $X := x_1, \LL, x_n$
where $n \geq 0$, with respective domains ${\cal D} := D_1, \LL, D_n$
associated with them.  So each variable $x_i$ ranges over the domain
$D_i$.  By a {\em constraint} $C$ on $X$ we mean a subset of $D_1
\times \LL \times D_n$.  
In this paper we consider only finite domains.

By a {\em constraint satisfaction problem}, in short CSP, we mean a
finite sequence of variables $X$ with respective domains ${\cal
  D}$, together with a finite set $\cal C$ of constraints, each on a
subsequence of $X$. We write it as $\p{{\cal C}}{x_1 \in D_1,
  \LL, x_n \in D_n}$, where $X := x_1, \LL, x_n$ and ${\cal D} :=
D_1, \LL, D_n$.

Consider now an element $d := d_1, \LL, d_n$ of $D_1 \times \LL \times
D_n$ and a subsequence $Y := x_{i_1}, \LL, x_{i_\ell}$ of
$X$. Then we denote by $d[Y]$ the sequence
$d_{i_1}, \LL, d_{i_{\ell}}$.

By a {\em solution\/} to  $\p{{\cal C}}{x_1 \in D_1, \LL, x_n \in D_n}$
we mean an element $d \in D_1 \times \LL \times D_n$ such that for
each constraint $C \in {\cal C}$ on a sequence of variables $X$
we have $d[X] \in C$.
We call a CSP {\em consistent\/}
if it has a solution.

Consider now a constraint $C$ on a sequence of variables $X$.
Given a subsequence $Y$ of $X$ by the {\em domain of $Y$\/} we
mean the set of all tuples from  $D_1 \times \LL \times D_k$, where
$D_1, \LL, D_k$ are the respective domains of the variables from $Y$.

In the introduction we informally referred to the notion of a CSP
``being built out of predefined, explicitly given finite
constraints.''  Let us make now this concept formal.
We need two auxiliary notions first, where in preparation for the
next definition we already consider constraints together with the
domains over which they are defined.

\begin{definition} 
\mbox{}
  \begin{itemize}
  \item Given a constraint $C \sse D_1 \times \LL \times D_n$
and a permutation $\pi$ of $[1..n]$ we denote by
$C^{\pi}$ the relation defined by
\[
(a_1, \LL, a_n) \in C^{\pi} \mbox{ iff } (a_{\pi(1)}, \LL, a_{\pi(n)}) \in C.
\]

\item Given two constraints 
$C \sse D_1 \times \LL \times D_n$ and $E \sse D'_1 \times \LL \times D'_n$
we say that {\em $C$ is based on $E$\/} if 
\begin{itemize}
\item $D_i \sse D'_i$ for $i \in [1..n]$,

\item $C = E \cap (D_1 \times \LL \times D_n)$.
\HB
\end{itemize}
  \end{itemize}
\end{definition}
So the notion of ``being based on'' involves
the domains of both constraints.
If $C$ is based on $E$, then $C$ is the restriction of
$E$ to the domains over which $C$ is defined.

\begin{definition} \label{def:based}

We assume that the ``predefined constraints'' are presented as a given
in advance CSP ${\cal BASE}$ and the considered CSP ${\cal P}$ is
related to ${\cal BASE}$ as follows:

\begin{itemize}

\item There is a mapping $f$ that relates 
each constraint $C$ of ${\cal P}$ to a 
constraint $f(C)$ of ${\cal BASE}$.

\item Each constraint $C$ of ${\cal P}$ is based on
$f(C)^{\pi}$, where $\pi$ is a permutation of $[1..n]$
and $n$ the arity of $C$.
\end{itemize}

We say then that ${\cal P}$ is {\em based on\/}
${\cal BASE}$.
\HB
\end{definition}

In the above definition the ``permuted'' relations $R^{\pi}$ allow us
to abstract from the variable ordering used in ${\cal BASE}$.
The following example illustrates this notion.

\begin{example} \label{exa:full-adder}
Consider the well-known full adder circuit.
It is defined by the following formula:
\[
\begin{array}{l}
add(i_1,i_2,i_3,o_1,o_2) \equiv \\
~~        xor(i_1,i_2,x_1),
        and(i_1,i_2,a_1),
        xor(x_1,i_3,o_2),
        and(i_3,x_1,a_2),
        or(a_1,a_2,o_1),
\end{array}
\]
where $and, xor$ and $or$ are defined in the expected way.
We can view the original constraints as the following CSP:
\[
{\cal BOOL} := \p{and(x,y,z),  xor(x,y,z),  or(x,y,z)}
                 {x \in \C{0,1}, y \in \C{0,1}, z \in \C{0,1}}.
\]

${\cal BOOL}$ should be viewed just as an ``inventory'' of the
predefined constraints and not as a CSP to be solved.  Now, any query
concerning the full adder can be viewed as a CSP based on ${\cal
BOOL}$.  For example, in Section \ref{sec:examples} we shall consider
the query $add(1,x,y,z,0)$. It corresponds to the following CSP based
on ${\cal BOOL}$:
\begin{tabbing}
\ $\langle$ \= $xor(i_1,i_2,x_1), \ and(i_1,i_2,a_1), \ xor(x_1,i_3,o_2), \ and(i_3,x_1,a_2), \ or(a_1,a_2,o_1)\ ;$ \\
\> $i_1 \in \C{1}, 
                        i_2 \in \C{0,1}, 
                        i_3 \in \C{0,1}, 
                        o_1 \in \C{0,1}, 
                        o_2 \in \C{0}, 
                        a_1 \in \C{0,1}, 
                        a_2 \in \C{0,1},$ \\
\>  $x_1 \in \C{0,1} ~ \rangle$.
\end{tabbing}
\HB
\end{example}

\section{Rule Consistency}
\label{sec:rule-consistency}

Our considerations crucially rely on the following notion of a rule.

\begin{definition}
Consider a constraint $C$ on a sequence of variables {\it VAR\/},
a subsequence $X$ of {\it VAR\/} and a variable $y$ of {\it VAR\/} not
in $X$, a tuple $s$ of elements
from the domain of $X$ and an element $a$ from the domain of $y$.
We call $X = s \ra y \neq a$ a {\em rule\/} ({\em for\/} $C$). 

\begin{itemize}
\item 
We say that 
$X = s \ra y \neq a$ is {\em valid\/} ({\em for\/} $C$)
if for every tuple $d \in C$ the equality $d[X] = s$ implies $d[y] \neq a$.

\item We say that 
$X = s \ra y \neq a$ is {\em feasible\/} ({\em for\/} $C$)
if for some tuple $d \in C$ the equality $d[X] = s$ holds.

\item 

Suppose that
$X := x_{1}, \LL, x_{k}$ and $s := s_1, \LL, s_k$.
We say that $C$ is {\em closed under the rule $X = s \ra y \neq a$}
if the fact that the domain of each variable $x_{j}$ equals $\C{s_j}$
implies that $a$ is not an element of the domain of the variable $y$.
\end{itemize}

Further, given a sequence of variables $Z$ that extends $X$ and a
tuple of elements $u$ from the domain of $Z$ that extends $s$, we say
that the rule $Z = u \ra y \neq a$ {\em extends\/} $X = s \ra y \neq
a$.  We call a rule {\em minimal\/} if it is feasible and it does not
properly extend a valid rule.  
\HB
\end{definition}

Note that rules that are not feasible are trivially valid.
To illustrate the introduced notions consider the following example.

\begin{example} \label{exa:conjunction}
  Take as a constraint the ternary relation that represents the
  conjunction $and(x, y, z)$. It can be viewed as the
  following relation: \noprint{table: 
\III

\begin{center}
\begin{tabular}{|l|l|l|}
\hline
$x$ & $y$ & $z$ \\ \hline \hline
0 & 0 & 0 \\ 
0 & 1 & 0 \\ 
1 & 0 & 0 \\ 
1 & 1 & 1 \\ \hline
\end{tabular}
\end{center}}
\[
\{(0,0,0),(0,1,0),(1,0,0),(1,1,1)\}.
\]
In other words, we assume that each of the variables $x,y,z$
has the domain $\C{0,1}$ and view
$and(x,y,z)$ as the constraint on $x,y,z$ that consists of
the above four triples.

It is easy to see that the rule $x = 0 \ra z \neq 1$ is valid for
$and(x,y,z)$. Further, the rule $x = 0, y = 1 \ra z \neq 1$ extends
the rule $x = 0 \ra z \neq 1$ and is also valid for
$and(x,y,z)$. However, out of these two rules only $x = 0 \ra z \neq
1$ is minimal.

Finally, both rules are feasible while the rules
$x = 0, z = 1 \ra y \neq 0$ and $x = 0, z = 1 \ra y \neq 1$ 
are not feasible.
\HB
\end{example}

\noprint{
Note that the rules just introduced are more expressive than so-called
{\em dependency rules\/} of database systems (see,
e.g. \cite{Ull88}). The dependency rules are of the form $X = s
\ra y = a$ with an obvious interpretation. They can be modelled by
means of the rules here considered.  Indeed, each such rule is
equivalent to the conjunction of the rules of the form $X = s \ra y
\neq b$ with $b \in D -\C{a}$, where $D$ is the domain of the variable
$y$.

However, the converse is not true, as can be seen by taking the
variables $x,y$, each with the domain $\C{0,1,2}$, and the constraint
$C$ on $x,y$ represented by the following table: 
\III

\begin{center}
\begin{tabular}{|l|l|}
\hline
$x$ & $y$ \\ \hline \hline
0 & 1 \\ 
0 & 0 \\ 
2 & 2 \\ \hline
\end{tabular}
\end{center}
Then the rule $x = 0 \ra y \neq 2$ is valid for $C$ but it is not equivalent
to a conjunction of the dependency rules.
}

Note \noprint{also} that a rule that extends a valid rule is valid, as well.  So
validity extends ``upwards''.  

Next, we introduce a notion of local consistency that is expressed in terms
of rules.

\begin{definition} \label{def:rule-cons}
Consider a CSP ${\cal P}$ is based on a CSP ${\cal BASE}$.  Let
$C$ be a constraint of ${\cal P}$ on the variables $x_1, \LL, x_n$
with respective non-empty domains $D_1,$ $\LL, D_n$. For some constraint
$f(C)$ of ${\cal BASE}$ and a permutation $\pi$ we have
$C = f(C)^{\pi} \cap (D_1 \times \LL \times D_n)$.
 
\begin{itemize}
  
\item We call the constraint $C$ {\em rule consistent\/} (w.r.t.  ${\cal BASE}$)
if it is closed under all rules that are valid for $f(C)^{\pi}$.

\item We call a CSP {\em rule consistent\/} (w.r.t.  ${\cal BASE}$)
if all its constraints are rule consistent.
\end{itemize}
\HB
\end{definition}

In what follows we drop the reference to ${\cal BASE}$ if it is clear from 
the context.

\begin{example}
Take as the base CSP
\[
{\cal BASE} := \p{and(x,y,z)}{x \in \C{0,1}, y \in \C{0,1}, z \in \C{0,1}}
\]
and consider the following four CSP's based on it:
\begin{enumerate}

\item $\p{and(x,y,z)}{x \in \C{0}, y \in D_y, z \in \C{0}}$,

\item $\p{and(x,y,z)}{x \in \C{1}, y \in D_y, z \in \C{0,1}}$,

\item $\p{and(x,y,z)}{x \in \C{0,1}, y \in D_y, z \in \C{1}}$,

\item $\p{and(x,y,z)}{x \in \C{0}, y \in D_y, z \in \C{0,1}}$,

\end{enumerate}
where $D_y$ is a subset of $\C{0,1}$.  We noted in Example
\ref{exa:conjunction} that the rule $x = 0 \ra z \neq 1$ is valid for
$and(x,y,z)$.  In the first three CSP's its only constraint is closed
under this rule, while in the fourth one not since 1 is present in the
domain of $z$ whereas the domain of $x$ equals $\C{0}$.  So the fourth
CSP is not rule consistent.  One can show that the first two CSP's are
rule consistent, while the third one is not since it is not closed
under the valid rule $z = 1 \ra x \neq 0$.
\end{example}

The following observation is useful.

\begin{note} \label{not:rule-cons}
Consider two constraints $C$ and $E$ such that $C \sse E$. 
Then $C$ is closed under all valid rules for $E$ iff
it is closed under all minimal valid rules for $E$.
\end{note}
\noprint{\Proof 
Note that if $C$ is closed under a rule $r_1$ and the rule
$r_2$ extends $r_1$ then $C$ is also closed under $r_2$.
Additionally, every valid feasible rule for $E$ extends some minimal
valid rule for $E$.  Moreover, since $C \sse E$, the constraint $C$ is
trivially closed under all rules that are not feasible for $E$. 
\HB 
\VV}

This allows us to confine our attention to minimal valid rules.
We now introduce an algorithm that given a constraint generates the 
set of all minimal valid rules for it.  We collect the
generated rules in a list.  We denote below the empty list by {\bf empty}
and the result of insertion of an element $r$ into a list $L$ by ${\bf
  insert}(r, L)$.

By an {\em assignment\/} to a sequence
of variables $X$ we mean here an element $s$ from the domain of $X$
such that for some $d \in C$ we have $d[X] = s$.  Intuitively, if we
represent the constraint $C$ as a table with rows corresponding to the
elements (tuples) of $C$ and the columns corresponding to the
variables of $C$, then an assignment to $X$ is a tuple of elements
that appears in some row in the columns that correspond to the variables
of $X$. 
This algorithm has the following form where we assume that the
considered constraint $C$ is defined on a sequence of variables {\em
  VAR\/} of cardinality $n$.   \II

\NI
{\sc Rules Generation} algorithm
\noprint{}
\begin{alltt}
\emph{L := {\bf empty}};
FOR \emph{i:= 0} TO \emph{n-1} DO
   FOR each subset \emph{X} of \emph{VAR} of cardinality \emph{i} DO
      FOR each assignment \emph{s} to \emph{X} DO
         FOR each \emph{y} in \emph{VAR-X} DO
            FOR each element \emph{d} from the domain of \emph{y} DO
               \emph{r :=  X = s} \ra \emph{y} \(\neq\) \emph{d};
               IF \emph{r} is valid for \emph{C} 
                  and it does not extend an element of \emph{L} 
                  THEN \emph{{\bf insert}(r, L)}
\end{alltt}
\noprintpb{
               END
            END
         END
      END
   END
END
\end{alltt}
}

The following result establishes correctness of this algorithm.

\begin{theorem} \label{thm:rga}
Given a constraint $C$ the {\sc Rules Generation} algorithm produces
in {\it L} the set of all minimal valid rules for $C$.
\end{theorem}
\noprint{\Proof 
First note that in the algorithm all possible feasible rules are
considered and in the list {\it L} only the valid rules are retained.
Additionally, a valid rule is retained only if it does not extend a
rule already present in {\it L}. Because the rules are
considered in the order according to which those that use less
variables are considered first, this ensures that precisely
all minimal rules valid are retained in {\it L}.
\HB
\VV}

Note that because of the minimality property no rule
in $L$ extends another.


\section{Relating Rule Consistency to Arc Consistency}
\label{sec:relation-arc-consistency}

To clarify the status of rule consistency we compare it now
to the notion of arc consistency. This notion was introduced in
\cite{mackworth-consistency} for binary relations and was extended to
arbitrary relations in  \cite{MM88}. Let us recall the
definition.

\begin{definition} \mbox{} \\[-6mm]
  \begin{itemize}
  \item We call a constraint $C$ on a sequence of variables $X$ {\em arc
      consistent\/} if for every variable $x$ in $X$ and an
    element $a$ in its domain there exists $d \in C$ such that $a =
    d[x]$.  That is, each element in each
    domain participates in a solution to $C$.

\item We call a CSP {\em arc consistent\/} if all its
constraints are arc consistent.
\HB
  \end{itemize}
\end{definition}

The following result relates  for constraints of arbitrary 
arity arc consistency to rule consistency.

\begin{theorem} \label{thm:rulecons}
Consider a CSP $\cal P$ based on a CSP ${\cal BASE}$.
If $\cal P$ is arc consistent then it is rule consistent
w.r.t. ${\cal BASE}$.
\end{theorem}
\noprint{\Proof 
Assume that $\cal P$ is arc consistent.  Choose a constraint $C$ of
$\cal P$ and consider a rule $X = s \ra y \neq a$ that is valid for
$f(C)^{\pi}$, where $f$ and $\pi$ are as in Definition
\ref{def:based}.

Suppose by contradiction that $C$ is not closed
under this rule. So for $X := x_{1}, \LL, x_{k}$ and $s := s_1, \LL,
s_k$ the domain of each variable $x_{j}$ in $\cal P$ equals $\C{s_j}$
and moreover $a \in D$, where $D$ is the domain of the variable $y$ in
$\cal P$.

By the arc consistency of $\cal P$ there exists $d \in C$ such that
$d[y] = a$. Because of the form of the domains of the variables in
$X$, also $d[X] = s$ holds.  Additionally, because $\cal P$ is based
on ${\cal BASE}$, we have $d \in f(C)^{\pi}$.  But by assumption
the rule $X = s \ra y \neq a$ is valid for $f(C)^{\pi}$, so $d[y] \neq
a$.  A contradiction.  
\HB 
\VV}

The converse implication does not hold in general as the following
example shows.

\begin{example} \label{exa:not-arc-consistent}
Take as the base the following CSP
\[
{\cal BASE} := \p{C}{x \in \C{0,1,2}, y \in \C{0,1,2}}
\]
where the constraint $C$ on $x,y$ that equals the set $\C{(0,1), (1,0), (2,2)}$.
So $C$ can be viewed as the following relation: \noprint{table:
\III

\begin{center}
\begin{tabular}{|l|l|}
\hline
$x$ & $y$ \\ \hline \hline
0 & 1 \\ 
1 & 0 \\ 
2 & 2 \\ \hline
\end{tabular}
\end{center}}
\[
\{(0,1), (1,0), (2,2)\}.
\]
Next, take for $D_1$ the set $\C{0,1}$ and $D_2$ the set $\C{0,1,2}$.
Then the CSP $\p{C \cap (D_1 \times D_2)}{x \in D_1, y \in D_2}$, so
$\p{\C{(0,1), (1,0)}}{x \in \C{0,1}, y \in \C{0,1,2}}$ is based on
${\cal BASE}$ but is not arc consistent since the value 2 in the
domain of $y$ does not participate in any solution. Yet, it is easy to
show that the only constraint of this CSP is closed under all rules
that are valid for $C$.  
\HB
\end{example}

However, if each domain has at most two elements, then the
notions of arc consistency and rule consistency coincide.  More
precisely, the following result holds.

\begin{theorem} \label{thm:rule-consistency}
  Let ${\cal BASE}$ be a CSP each domain of which is unary or binary.
  Consider a CSP $\cal P$ based on ${\cal BASE}$.  Then $\cal P$ is
  arc consistent iff it is rule consistent w.r.t. ${\cal BASE}$.
\end{theorem}

\section{Inclusion Rule Consistency}
\label{sec:inclusion-rule-consistency}

We saw in the previous section that the notion of rule consistency is weaker
than that of arc consistency for constraints of arbitrary arity. 
We now show how by modifying the format
of the rules we can achieve arc consistency.
To this end we introduce the following notions.

\begin{definition}
Consider a constraint $C$ over a sequence variables {\it VAR\/},
a subsequence $X := x_1, \LL, x_k$ of {\it VAR\/} and a variable $y$ of {\it VAR\/} not
in $X$, a tuple $S := S_1, \LL, S_k$ of respective subsets of
the domains of the variables from $X$ and an element $a$ from the domain of $y$.

We call $X \sse S \ra y \neq a$ an {\em inclusion rule\/} ({\em for\/}
$C$).  We say that $X \sse S \ra y \neq a$ is {\em valid\/} ({\em
  for\/} $C$) if for every tuple $d \in C$ the fact that $d[x_i] \in
S_i$ for $i \in [1..k]$ implies that $d[y] \neq a$ and that $X \sse S \ra
y \neq a$ is {\em feasible\/} ({\em for\/} $C$) if for some tuple $d
\in C$ we have $d[x_i] \in S_i$ for $i \in [1..k]$.

Further, we say that a constraint $C$ is {\em closed under the inclusion rule
  $X \sse S$ $\ra y \neq a$} if the fact that the domain of each
variable $x_{j}$ is included in $S_j$ implies that $a$ is not an
element of the domain of the variable $y$.  
\HB
\end{definition}

By choosing in the above definition singleton sets  $S_1, \LL, S_k$ we see
that the inclusion rules  generalize the rules of Section
\ref{sec:rule-consistency}.  
Note that inclusion rules that are not feasible are trivially valid.

In analogy to Definition \ref{def:rule-cons} we now introduce the following notion.

\begin{definition} \label{def:inclusion-rule-cons}
Consider a CSP ${\cal P}$ is based on a CSP ${\cal BASE}$.  Let
$C$ be a constraint of ${\cal P}$ on the variables $x_1, \LL, x_n$
with respective non-empty domains $D_1,$ $\LL,$$D_n$. For some constraint
$f(C)$ of  ${\cal BASE}$ and a permutation $\pi$ we have
$C = f(C)^{\pi} \cap (D_1 \times \LL \times D_n)$.
 
\begin{itemize}
  
\item We call the constraint $C$ {\em inclusion rule consistent\/} (w.r.t.  ${\cal BASE}$)
if it is closed under all inclusion rules that are valid for $f(C)^{\pi}$.

\item We call a CSP {\em inclusion rule consistent\/} (w.r.t.  ${\cal BASE}$)
if all its constraints are inclusion rule consistent.
\HB
\end{itemize}
\end{definition}

We now have the following result.
\begin{theorem} \label{thm:arccons}
Consider a CSP $\cal P$ based on a CSP ${\cal BASE}$.
Then $\cal P$ is arc consistent iff it is 
inclusion rule consistent
w.r.t. ${\cal BASE}$.
\end{theorem}

Example \ref{exa:not-arc-consistent} shows that the notions of rule
consistency and inclusion rule consistency do not coincide. 
\noprint{To 
see this difference better let us reconsider the CSP discussed
in this example.

We noted there that this CSP is not arc consistent and that it
is rule consistent. From the above theorem we know that this CSP
is not inclusion rule consistent. In fact, consider the following
inclusion rule:
\[
x \subseteq \C{0,1} \ra y \neq 2.
\]

This inclusion rule is valid for the base constraint $C$ but the
restricted constraint $C \cap (D_1 \times D_2)$ is not closed under
this rule.  In conclusion, the inclusion rules are more powerful than
the rules considered in Section \ref{sec:rule-consistency}.
}

In Section \ref{sec:rule-consistency} we introduced an algorithm that
given a constraint $C$ generated the set of all minimal rules valid
for $C$.  We now modify it to deal with the inclusion rules.  First we
need to adjust the notions of an extension and of minimality.

\begin{definition}
  Consider a constraint $C$ on a sequence of variables {\it VAR}.  Let
  $X := x_1, \LL, x_k$ and $Z := z_1, \LL, z_{\ell}$ be two
  subsequences of {\it VAR\/} such that $Z$ extends $X$ and $y$ a
  variable of {\it VAR\/} not in $Z$. Further, let $S := S_1, \LL,
  S_k$ be the sequence of respective subsets of the domains of the
  variables from $X$, $U := U_1, \LL, U_{\ell}$ the sequence of
  respective subsets of the domains of the variables from $Z$, and $a$
  an element from the domain of $y$.

  We say that the inclusion rule $\ r_1 := Z \sse U \ra y \neq a$ {\em
    extends\/} $\ r_2 := X \sse S \ra y \neq a$ if for each common
  variable of $X$ and $Z$ the corresponding element of $U$ is a subset
  of the corresponding element of $S$.  We call an inclusion rule {\em
    minimal\/} if it is feasible and it does not properly extend a
  valid inclusion rule.  
\HB
\end{definition}

To clarify these notions consider the following example.
\begin{example} \label{exa:waltz}
Consider a constraint on variables $x,y,z$, each with the 
domain $\C{+,-, l, r}$, that is defined by the 
following relation: \noprint{table:
\III

\begin{center}
\begin{tabular}{|l|l|l|}
\hline
$x$ & $y$ & $z$ \\ \hline \hline
$+$ & $+$ & $+$ \\
$-$ & $-$ & $-$ \\
$l$ & $r$ & $-$ \\
$-$ & $l$ & $r$ \\
$r$ & $-$ & $l$ \\ \hline
\end{tabular}
\end{center}}
\[
\{(+,+,+),(-,-,-),(l,r,-),(-,l,r),(r,-,l)\}
\]
This constraint is the so-called {\em fork\/} junction
in the language of \cite{waltz75} 
for describing polyhedreal scenes.
Note that the following three inclusion rules
\[
r_1 := 
x \subseteq \C{+,-} \ra z \neq l,
\]
\[
r_2 := x \subseteq \C{+} \ra z \neq l,
\]
and
\[
r_3 := x \subseteq \C{-}, y \sse \C{l} \ra z \neq l
\] 
are all valid.  Then the inclusion rules $r_2$ and $r_3$
extend $r_1$ while the inclusion rule $r_1$ extends neither
$r_2$ nor $r_1$. Further,
the inclusion rules $r_2$ and $r_3$ are incomparable in the sense
that none extends the other.  
\HB
\end{example}

The following counterpart of Note \ref{not:rule-cons} holds.

\begin{note} \label{not:inclusion-rule-cons}
Consider two constraints $C$ and $E$ such that $C \sse E$. 
Then $C$ is closed under all valid inclusion rules for $E$ iff
it is closed under all minimal valid inclusion rules for $E$.
\end{note}




As in Section \ref{sec:rule-consistency} we now provide an algorithm
that given a constraint generates the set of all minimal valid
inclusion rules.
We assume here that the considered constraint $C$ is defined on a
sequence of variables {\em VAR\/} of cardinality $n$.  

Instead of assignments that are used in the {\sc Rules Generation}
algorithm we now need a slightly different notion. To define it for
each variable $x$ from {\em VAR\/} we denote the set $\C{d[x] \mid d
  \in C}$ by $C[x]$.  By a {\em weak assignment\/} to a sequence of
variables $X := x_1, \LL, x_k$ we mean here a sequence $S_1, \LL, S_k$
of subsets of, respectively, $C[x_1], \LL, C[x_k]$ such that some $d
\in C$ exists such that $d[x_i] \in S_i$ for each $i \in [1..k]$.

Intuitively, if we represent the constraint $C$ as a table with rows
corresponding to the elements of $C$ and the columns corresponding to
the variables of $C$ and we view each column as a set of elements,
then a weak assignment to $X$ is a tuple of subsets of the columns
that correspond to the variables of $X$ that ``shares'' an assignment.

In the algorithm below the weak assignments to a fixed sequence of
variables are considered in decreasing order in the sense that if the
weak assignments $S_1, \LL, S_k$ and $U_1, \LL, U_k$ are such that for
$i \in [1..k]$ we have $U_i \sse S_i$, then $S_1, \LL, S_k$ is
considered first.

\III

\NI
{\sc Inclusion Rules Generation} algorithm

\begin{alltt}
\emph{L := {\bf empty}};
FOR \emph{i:= 0} TO \emph{n-1} DO
  FOR each subset \emph{X} of \emph{VAR} of cardinality \emph{i} DO
    FOR each weak assignment \emph{S} to \emph{X} in decreasing order DO 
      FOR each \emph{y} in \emph{VAR-X} DO
        FOR each element \emph{d} from the domain of \emph{y} DO
           \emph{r :=  X} \(\subseteq\) \emph{S} \ra \emph{y} \(\neq\) \emph{d};
           IF \emph{r} is valid for \emph{C} 
              and it does not extend an element of \emph{L}
              THEN \emph{{\bf insert}(r, L)}
\end{alltt}
\noprintpb{
               END
            END
         END
      END
   END
END
\end{alltt}
}

The following result establishes correctness of this algorithm.

\begin{theorem} \label{thm:irga}
Given a constraint $C$ the {\sc Inclusion Rules Generation} algorithm produces
in {\it L} the set of all minimal valid inclusion rules for $C$.
\end{theorem}

\section{Applications}
\label{sec:examples}
In this section we discuss the implemention of the 
{\sc Rules Generation} and {\sc Inclusion Rules Generation} algorithms
and discuss their use on selected domains.

\subsection{Constraint Handling Rules ({\tt CHR})}

In order to validate our approach we have realized in the Prolog
platform \eclipse{} a prototype implementation of both the
\textsc{Rules Generation} algorithm and the \textsc{Inclusion Rules
  Generation} algorithm.  
\noprint{We made a compromise between memory usage
and performance so that we could tackle some non-trivial problems (in
terms of size of the domains of variables, and in terms of arity of
constraints) in spite of the exponential complexity of the algorithms.}
These implementations generate {\tt CHR} rules that deal with finite
domain variables using an \eclipse{} library.

Constraint Handling Rules ({\tt CHR}) of \cite{fruhwirth-constraint-95}
is a declarative language that
allows one to write guarded rules for rewriting constraints. These
rules are repeatedly applied until a fixpoint is reached. The rule
applications have a precedence over the usual resolution step of logic
programming.

{\tt CHR} provides two types of rules: simplification rules that replace a
constraint by a simpler one, and propagation rules that add new
constraints.  

Our rules and inclusion rules can be modelled by means of propagation
rules. To illustrate this point consider some constraint $cons$
on three variables, $A,B,C$, each with the domain $\C{0,1,2}$.

The \textsc{Rules Generation} algorithm generates rules such as
$(A,C)=(0,1) \rightarrow B \neq 2$. This rule is translated into a
{\tt CHR} rule of the form: \verb+cons(0,B,1)+ \verb+==>+ \verb+B##2+.
Now, when a constraint in the program query matches
\verb+cons(0,B,1)+, this rule is fired and the value 2 is removed from
the domain of the variable \verb+B+.

In turn, the \textsc{Inclusion Rules Generation} algorithm generates
rules such as $(A,C) \subseteq (\{0\},\{1,2\}) \rightarrow B \neq 2$.
This rule is translated into the {\tt CHR} rule
\begin{verbatim}
cons(0,B,C) ==>in(C,[1,2]) | B##2
\end{verbatim}
where the {\tt in} predicate is defined by
\begin{verbatim}
in(X,L):- dom(X,D), subset(D,L).
\end{verbatim}
So {\tt in(X,L)} holds if the current domain of the variable {\tt
X} (yielded by the built-in {\tt dom} of \eclipse{}) is included in
the list {\tt L}.

Now, when a constraint matches \verb+cons(0,B,C)+ {\em and\/} the
current domain of the variable \verb+C+ is included in \verb+[1,2]+,
the value 2 is removed from the domain of \verb+B+.
So for both types of rules we achieve the desired effect.

In the examples below we combine the rules with the same premise into
one rule in an obvious way and present these rules in the {\tt CHR}
syntax.

\subsection{Generating the rules}

We begin by discussing the generation of rules and inclusion rules
for some selected domains.
The times given refer to an implementation ran on a Silicon Graphics O2 
with 64 Mbytes of memory and a 180 MHZ processor.

\paragraph{Boolean constraints}
As the first example consider the Boolean constraints, for
example the conjunction constraint {\tt and(X,Y,Z)} 
of Example \ref{exa:conjunction}.
The {\sc Rules Generation} algorithm
generated in 0.02 seconds the following six rules:
\begin{verbatim}
and(1,1,X) ==> X##0.
and(X,0,Y) ==> Y##1.
and(0,X,Y) ==> Y##1.
and(X,Y,1) ==> X##0,Y##0.
and(1,X,0) ==> X##1.
and(X,1,0) ==> X##1.
\end{verbatim}

Because the domains are here binary we can 
replace the conclusions of the form {\tt U \#\# 0} by {\tt U = 1} 
and {\tt U \#\# 1} by {\tt U = 0}.
These become then the well-known rules that can be found in
\cite[page 113]{FruehwirthJLP98}. 

In this case, by virtue of Theorem \ref{thm:rule-consistency}, the
notions of rule and arc consistency coincide, so the above six rules 
characterize the arc consistency of the {\tt and} constraint. Our 
implementations of the {\sc Rules Generation} and the {\sc Inclusion 
Rules Generation} algorithms yield here the same rules.

\paragraph{Three valued logics}

Next,  consider the three valued logic of 
\cite[page 334]{Kle52} that consists of three values,
t (true), f (false) and u (unknown).
We only consider here the crucial equivalence relation $\equiv$ defined by
the truth table

\begin{center}
\begin{tabular}{|c|ccc|} \hline
$\equiv$ & t & f & u \\ \hline
t & t & f & u \\
f & f & t & u \\
u & u & u & u \\
\hline
\end{tabular}
\end{center}
that determines a ternary constraint with nine triples.
We obtain for it 20 rules and 26 inclusion rules.
Typical examples are
\begin{alltt}
equiv(X,Y,f) ==> X##u,Y##u.
\textrm{and}
equiv(t,X,Y) ==> in(Y,[f, u]) | X##t.
\end{alltt}

\noprint{
\paragraph{Propagating signs}

As a next example consider the rules for propagating signs
in arithmetic expressions, see, e.g. 
\cite[page 303]{davis87}. We limit ourselves to the
case of multiplication.
Consider the following table:
\III

\begin{center}
\begin{tabular}{|r|llll|} \hline
$\times$ & neg  & zero  & pos   & unk \\
\hline
neg     & pos   & zero  & neg   & unk \\
zero    & zero  & zero  & zero  & zero \\
pos     & neg   & zero  & pos   & unk \\
unk     & unk   & zero  & unk   & unk \\
\hline
\end{tabular}
\end{center}

\NI
This table determines a ternary constraint {\tt msign}
that consists of 16 triples, for
instance (neg, neg, pos), that denotes the fact that the
multiplication of two negative numbers yields a positive number.
The value ``unk'' stands for ``unknown''.
The {\sc Rules Generation} algorithm
generated in 0.08 seconds 34 rules.
A typical example is \verb+msign(X,zero,Y) ==> Y##pos,Y##neg,Y##unk+.

In turn, the {\sc Inclusion Rules Generation} algorithm
generated in 0.6 seconds 54 inclusion rules. A typical example is
\begin{verbatim}
msign(X,unk,Y) ==> in(Y,[neg, pos, zero]) | X##pos,X##neg
\end{verbatim}
that corresponds to the following two rules (for the constraint {\tt msign(X,Z,Y)}) in the notation of Section
\ref{sec:inclusion-rule-consistency}: 
\[Z \sse \C{{\tt unk}}, Y \sse \C{{\tt neg, pos, zero}} \ra X \neq {\tt pos}
\]
and
\[
Z \sse \C{{\tt unk}}, Y \sse \C{{\tt neg, pos, zero}} \ra X \neq {\tt neg}.
\]
}

\paragraph{Waltz' language for describing polyhedreal scenes}

Waltz' language consists of four constraints. One of them, the fork
junction was already mentioned in Example \ref{exa:waltz}.  The {\sc
  Rules Generation} algorithm generated for it 12
rules and the {\sc Inclusion Rules Generation} algorithm 24 inclusion
rules.

Another constraint, the so-called T junction, is defined by
the following relation: \noprint{table:

\begin{center}
\begin{tabular}{|l|l|l|}
\hline
$x$ & $y$ & $z$ \\ \hline \hline
$r$ & $l$ & $+$ \\
$r$ & $l$ & $-$ \\
$r$ & $l$ & $r$ \\
$r$ & $l$ & $l$ \\ \hline
\end{tabular}
\end{center}}
\[
\{(r,l,+),(r,l,-),(r,l,r),(r,l,l)\}.
\]
In this case the {\sc Rules Generation} algorithm and the {\sc
  Inclusion Rules Generation} algorithm both generate the same output
that consists of just one rule:
\begin{verbatim}
t(X,Y,Z) ==> X##'l',X##'-',X##'+',Y##'r',Y##'-',Y##'+'.
\end{verbatim}
So this rule characterizes both rule consistency
and arc consistency for the CSP's based on the T junction.

For the other two constraints, the L junction and the arrow junction,
the generation of the rules and inclusion rules is equally 
straightforward.

\subsection{Using the rules}

Next, we show by means of some examples how the generated rules can be
used to reduce or to solve specific queries. Also, we show how using
compound constraints we can achieve local consistency notions
that are stronger than arc consistency for constraints of arbitrary arity.

\paragraph{Waltz' language for describing polyhedreal scenes}

The following predicate describes the impossible object given in
Figure~12.18 of \cite[page 262]{Win92}:
\begin{verbatim}
imp(AF,AI,AB,IJ,IH,JH,GH,GC,GE,EF,ED,CD,CB):-
     S1=[AF,AI,AB,IJ,IH,JH,GH,GC,GE,EF,ED,CD,CB],
     S2=[FA,IA,BA,JI,HI,HJ,HG,CG,EG,FE,DE,DC,BC],
     append(S1,S2,S), S :: [+,-,l,r],
      
     arrow(AF,AB,AI), l(BC,BA), arrow(CB,CD,CG),  
     l(DE,DC), arrow(ED,EG,EF), l(FA,FE), fork(GH,GC,GE), 
     arrow(HG,HI,HJ), fork(IA,IJ,IH), l(JH,JI),

     line(AF,FA), line(AB,BA), line(AI,IA), line(IJ,JI),
     line(IH,HI), line(JH,HJ), line(GH,HG), line(FE,EF), 
     line(GE,EG), line(GC,CG), line(DC,CD), line(ED,DE), 
     line(BC,CB).
\end{verbatim}
where the supplementary constraint {\tt line} is defined by the 
following relation: \noprint{table:

\begin{center}
\begin{tabular}{|l|l|}
\hline
$x$ & $y$  \\ \hline \hline
{\tt +} & {\tt +} \\
{\tt -} & {\tt -} \\
{\tt l} & {\tt r} \\
{\tt r} & {\tt l} \\ \hline
\end{tabular}
\end{center}}
\[
\{(+,+),(-,-),(l,r),(r,l)\}
\]
When using the rules obtained by the {\sc Rules Generation}
algorithm and associated with the \verb+fork+, \verb+arrow+, \verb+t+,
\verb+l+, and \verb+line+ constraints, the query
\begin{verbatim}
imp(AF,AI,AB,IJ,IH,JH,GH,GC,GE,EF,ED,CD,CB)
\end{verbatim} 
reduces in 0.009 seconds the variable domains to
\texttt{AF $\in$ [+,-, l]}, 
\texttt{AI $\in$ [+,-]}, 
\texttt{AB $\in$ [+,-,r]}, 
\texttt{IJ $\in$ [+,-,l,r]}, 
\texttt{IH $\in$ [+,-,l,r]}, 
\texttt{JH $\in$ [+,-,l,r]}, \\
\texttt{GH $\in$ [+,-,l,r]}, 
\texttt{GC $\in$ [+,-,l,r]}, 
\texttt{GE $\in$ [+,-,l,r]}, 
\texttt{EF $\in$ [+,-]},\\
\texttt{ED $\in$ [+,-,l]}, 
\texttt{CD $\in$ [+,-,r],} and
\texttt{CB $\in$ [+,-,l]}.

But
some constraints remain unsolved, so we need to add a labeling
mechanism to prove the inconsistency of the problem.  On the other
hand, when using the inclusion rules, the inconsistency is detected
without any labeling in 0.06 seconds.

In the well-known example of the cube given in Figure~12.15 of
\cite[page 260]{Win92} the inclusion rules are also more powerful
than the rules and both sets of rules reduce the problem but in both
cases labeling is needed to produce all four solutions.

\noprint{
Comparing the constraint solver based on the inclusion rules to the
constraint solver based on the rules is not easy: although propagation is
more efficient with the inclusion rules, the solver based on rules can
sometimes be faster depending on the structure of the problem and
on whether the labeling is needed.

We also compared the solvers generated by the implementations of 
the \textsc{Rules Generation} and \textsc{Inclusion Rules Generation} 
algorithms to
the approach described in~\cite{By96} and based on
meta-programming in Prolog. We ran the same examples and drew the
following conclusions.  For small examples our
solvers were less efficient than the ones of By (with factors varying
from 2 to 10).
However, for more complex examples, our solvers became
significantly more efficient, with factors varying from 10 to 500.
This can be attributed to the increased role of the constraint propagation
that reduces backtracking and that is absent in By's approach.
}

\paragraph{Temporal reasoning}

In \cite{All83} approach to temporal reasoning the entities are
intervals and the relations are temporal binary relations between
them.  \cite{All83} found that there are 13 possible temporal
relations between a pair of events, namely {\tt before, during,
  overlaps, meets, starts, finishes}, the symmetric relations of these
six relations and {\tt equal}. We denote these 13 relations respectively 
by {\tt b}, {\tt d}, {\tt o}, {\tt m}, {\tt s}, {\tt f}, {\tt b-},
{\tt d-}, {\tt o-},
{\tt m-}, {\tt s-}, {\tt f-}, {\tt e} and their set by {\tt TEMP}.

Consider now three events, {\tt A, B} and {\tt C} and suppose that we
know the temporal relations between the pairs {\tt A} and {\tt B}, and
{\tt B} and {\tt C}. The question is what is the temporal relation
between {\tt A} and {\tt C}.  To answer it \cite{All83} provided
a 13 $\times$ 13 table.  This table determines a ternary constraint
between a triple of events, {\tt A, B} and {\tt C} that we denote by
{\tt tr}.  For example,
\[
{\tt (overlaps, before, before)} \in {\tt tr}
\]
since {\tt A} {\tt overlaps\/}
{\tt B} and {\tt B} is {\tt before\/} {\tt C}
implies that  {\tt A} is {\tt before\/} {\tt C}.

Using this table, the {\sc Rule Generation}
algorithm produced for the constraint {\tt tr} 498 rules in 31.16
seconds.

We tried this set of rules to solve the following problem
from~\cite{All83}: ``John was not in the room when I touched the
switch to turn on the light.''. We have here three events: {\tt S}, the
time of touching the switch; {\tt L}, the time the light was on; and {\tt J},
the time that John was in the room.  Further, we have two relations:
{\tt R1} between {\tt L} and {\tt S}, and {\tt R2} between {\tt S} and 
{\tt J}. This problem
is translated into the CSP  $\p{{\tt tr}}{{\tt R1} \in {\tt [o-,m-]}, 
{\tt  R2} \in {\tt [b,m,b-,m-]}, {\tt R3} \in {\tt TEMP}}$, 
where {\tt tr} is the above constraint 
on the variables {\tt R1}, {\tt R2}, {\tt R3}.

To infer the relation {\tt R3} between {\tt L} and {\tt J} we can use the
following query\,\footnote{Since no variable is instantiated, we need
  to perform labeling to effectively apply the rules.}:
\begin{verbatim}
  R1::[o-,m-], R2::[b,m,b-,m-],
  R3::[b,d,o,m,s,f,b-,d-,o-,m-,s-,f-,e],
  tr(R1,R2,R3), labeling([R1,R2,R3]).
\end{verbatim}
We then obtain the following solutions in 0.06 seconds:
\texttt{(R1,R2,R3)} $\in$ 
\texttt{\{(m-,b,b),}
\texttt{(m-,b,d-),}
\texttt{(m-,b,f-),}
\texttt{(m-,b,m),}
\texttt{(m-,b,o),}
\texttt{(m-,b-,b-),}\\
\texttt{(m-,m,e),}
\texttt{(m-,m,s),} 
\texttt{(m-,m,s-),} 
\texttt{(m-,m-,b-),}
\texttt{(o-,b,b),}\\
\texttt{(o-,b,d-),}  
\texttt{(o-,b,f-),}
\texttt{(o-,b,m),} 
\texttt{(o-,b,o),} 
\texttt{(o-,b-,b-),}\\  
\texttt{(o-,m,d-),}
\texttt{(o-,m,f-),} 
\texttt{(o-,m,o),}
\texttt{(o-,m-,b-)\}}.

To carry on (as in~\cite{All83}), we now complete the problem
with: ``But John was in the room later while the light went out.''.
This is translated into: ``{\tt L} {\tt overlaps}, {\tt starts}, or 
{\tt is during} {\tt J}'',
i.e., {\tt R3} $\in$ {\tt [o,s,d]}.  

We now run the following query:
\begin{verbatim}
  R1::[o-,m-], R2::[b,m,b-,m-], R3::[o,s,d],
  tr(R1,R2,R3), labeling([R1,R2,R3]).
\end{verbatim}
and obtain four solutions in 0.04 seconds: \\
\texttt{(R1,R2,R3)} $\in$ 
\texttt{\{(m-,b,o),}
\texttt{(m-,m,s),}
\texttt{(o-,b,o),}
\texttt{(o-,m,o)\}}.

\paragraph{Full adder}
This final example illustrates how we can use the rules
and the inclusion rules to implement more powerful notions of local 
consistency.
The already discussed in Example \ref{exa:full-adder}
full adder circuit can be defined by the following
constraint logic program (see, e.g., \cite{FruehwirthJLP98})
that uses the Boolean constraints
{\tt and}, {\tt xor} and {\tt or}:
\begin{verbatim}
add(I1,I2,I3,O1,O2):-
        [I1,I2,I3,O1,O2,A1,A2,X1]:: 0..1,
        xor(I1,I2,X1), and(I1,I2,A1), xor(X1,I3,O2),
        and(I3,X1,A2), or(A1,A2,O1).
\end{verbatim}
The query \verb+add(I1,I2,I3,O1,O2)+ followed by a labeling mechanism
generates the explicit definition (truth table) of the {\tt
  full\_adder} constraint with eight entries such as
\verb+full_adder(1,0,1,1,0).+
\noprint{
\begin{verbatim}
full_adder(1, 0, 1, 1, 0).
\end{verbatim}
}

We can now generate rules and inclusion rules for the compound
constraint (here the {\tt full\_adder} constraint) that is defined by
means of some basic constraints (here the {\tt and, or} and {\tt xor}
constraints).  These rules refer to the compound constraint
and allow us to reason about it directly instead of by using the rules
that deal with the basic constraints.

In the case of the {\tt full\_adder} constraint the {\sc Rules
  Generation} algorithm generated 52 rules in 0.27 seconds.  The
constraint propagation carried out by means of these rules is more
powerful than the one carried out by means of the rules generated for
the {\tt and, or} and {\tt xor} constraints.\\
For example,
the query \verb+[X,Y,Z]::[0,1], full_adder(1,X,Y,Z,0)+ reduces {\tt Z} to
1 whereas the query
\verb+[X,Y,Z]::[0,1], add(1,X,Y,Z,0)+  does not reduce {\tt Z} at all.

This shows that the rule consistency for a compound constraint defined
by means of the basic constraints is in general stronger than the rule
consistency for the basic constraints treated separately.  In fact, in
the above case the
rules for the {\tt full\_adder} constraint yield the relational
(1,5)-consistency notion of \cite{DvB97}, whereas by virtue of
Theorem \ref{thm:rule-consistency}, the rules for the {\tt and, or} 
and {\tt xor} constraints yield a
weaker notion of arc consistency.

\section{Conclusions}

The aim of this paper was to show that constraint satisfaction
problems built out of explicitly given constraints defined over small
finite domains can be often solved by means of automatically generated
constraint propagation algorithms.

We argued that such CSP's often arise in practice and consequently
the methods here developed can be of practical use. Currently we
are investigating how the approach of this paper can be applied to a
study of various decision problems concerning specific multi-valued logics
and how this in turn could be used for an analysis of digital circuits.
Other applications we are now studying involve non-linear constraints
over small finite domains and the analysis of polyhedreal scenes in presence
of shadows (see \cite{waltz75}).

The introduced notion of rule consistency is weaker than arc
consistency and can be in some circumstances the more appropriate one
to use.  For example, for the case of temporal reasoning considered in
the last section we easily generated all 498 rules that enforce rule
consistency whereas 24 hours turned out not be enough to generate the
inclusion rules that enforce arc consistency.

Finally, the notions of rule consistency and inclusion rule
consistency could be parametrized by the desired maximal number of
variables used in the rule premises. Such parametrized versions of
these notions could be useful when dealing with constraints involving
a large number of variables.  Both the {\sc Rules Generation}
algorithm and the {\sc Inclusion Rules Generation} algorithm and their
implementations can be trivially adapted to such parametrized notions.

The approach proposed in this paper could be easily integrated into
constraint logic programming systems such as \eclipse{}. This could be
done by providing an automatic constraint propagation by means of the
rules or the inclusion rules for flagged predicates that are defined
by a list of ground facts, much in the same way as now constraint
propagation for linear constraints over finite systems is 
automatically provided.

\section*{Acknowledgements}
We would like to thank Thom Fr\"{u}hwirth, Andrea Schaerf and the
anonymous referees for useful suggestions concerning this paper.

\bibliographystyle{plain}


\end{document}